\documentclass[10pt,journal,compsoc]{IEEEtran}
\usepackage{amsmath,amssymb,amsfonts}
\usepackage{float}
\usepackage{multirow}
\usepackage{epsfig}
\usepackage{graphicx}
\usepackage{algorithm}
\usepackage{algorithmic}
\usepackage{subfigure}
\usepackage{epstopdf}
\usepackage{makecell}
\usepackage{epstopdf}
\usepackage{bm}
\usepackage{times}
\usepackage{booktabs}
\usepackage{microtype}
\usepackage{cite}
\usepackage{hyperref}
\usepackage{url}
\usepackage{color}

\ifCLASSINFOpdf
\else
\fi
\begin{document}
%
\title{Unified GCNs: Towards Connecting GCNs with CNNs}

\author{{Ziyan Zhang, Bo Jiang*, and Bin Luo,~\IEEEmembership{Member,~IEEE}} 
	\thanks{Ziyan Zhang, Bo Jiang and Bin Luo are
		with the Anhui Provincial Key Laboratory of Multimodal Cognitive Computation, School of Computer Science and Technology of Anhui University,
		Hefei, 230601, China.
		\protect\\
*Corresponding author: Bo Jiang (E-mail: jiangbo@ahu.edu.cn)}
\thanks{Manuscript received April 19, 2005; revised August 26, 2015.}}

\markboth{Journal of \LaTeX\ Class Files,~Vol.~14, No.~8, August~2015}%
{Shell \MakeLowercase{\textit{et al.}}: Bare Demo of IEEEtran.cls for Computer Society Journals}

\IEEEtitleabstractindextext{%
\begin{abstract}
Graph Convolutional Networks (GCNs)  have been widely demonstrated their powerful ability in graph data representation and learning.
Existing graph convolution layers are mainly designed based on graph signal processing and transform aspect
which usually suffer from some limitations, such as over-smoothing, over-squashing and non-robustness, etc. 
As we all know that Convolution Neural Networks (CNNs) have received great success in many computer vision and machine learning.
One main aspect is that CNNs leverage many learnable convolution filters (kernels) to obtain rich feature descriptors and thus can have high capacity to encode complex patterns in visual data analysis. Also, CNNs are flexible in designing their network architecture, such as MobileNet, ResNet, Xception, etc.
Therefore, it is natural to arise a question: \emph{can we design graph convolutional layer as flexibly as that in CNNs}?
Innovatively, in this paper, we consider connecting GCNs with CNNs deeply from a general perspective of depthwise separable convolution
operation.
Specifically, we show that GCN and GAT indeed perform some specific depthwise separable convolution operations.
 This novel interpretation enables us to better understand the
connections between GCNs (GCN, GAT) and CNNs and further inspires us to design more Unified GCNs (UGCNs).
As two showcases, we implement two UGCNs, i.e., Separable UGCN (S-UGCN) and General UGCN (G-UGCN) for graph data representation and learning.
Promising experiments on several graph representation benchmarks demonstrate the effectiveness and advantages of the proposed UGCNs.
\end{abstract}

\begin{IEEEkeywords}
Graph Convolutional Networks, Convolution Neural Networks, Depthwise Separable Convolution
\end{IEEEkeywords}}

\maketitle

\section{Introduction}

In many learning tasks, data is coming with graph form $G(X, A)$ where  ${X}\in \mathbb{R}^{N\times C}$ denotes node features and ${A}\in \mathbb{R}^{N\times N}$ represents edges encoding the relationships among nodes. 
Graph convolutional networks (GCNs)~\cite{spectralGCN,gcn,GCN_suervy3} have been demonstrated their powerful ability in dealing with graph data representation and learning tasks.
Existing graph convolutions (GCs) are mainly designed based on graph signal transform and processing perspective.
To be specific, existing GCs can be roughly categorized into two types, i.e., spectral GCs and spatial GCs.
Spectral methods aim to define GCs by employing graph signal processing/transform techniques in the spectral domain (\emph{e.g.}, Fourier domain), i.e., performing some frequency filtering on node's features $X$ of graph data~\cite{spectralGCN,GCN_suervy3}.
The first notable spectral GCN was proposed by Bruna et al.~\cite{spectralGCN} which
defines GC in the spectral space of graph Laplacian matrix.
Defferrard et al.~\cite{defferrard2016convolutional} propose to use $K$-polynomial filter for GC definition.
Kipf et al.~\cite{gcn} propose a simple Graph Convolutional Network (GCN) by exploring the first-order approximation of spectral filters.
Some more spectral GCs have also been developed~\cite{li2018adaptive,GWNN,Wang_2018_ECCV,zhu2021simple,levie2021transferability}.
Contrary to spectral method, spatial GCs are defined by employing graph signal processing and filtering techniques in the vertex/spatial domain directly.
For example,
Monti et al.~\cite{monti2017geometric} propose a general MoNet which designs node's representation by integrating the signals
within each node's neighbors.
Hamilton et al.~\cite{graphsage} present GraphSAGE for the general graph  inductive representation and learning tasks.
Veli{\v{c}}kovi{\'c} et al.~\cite{gat} propose Graph Attention Networks (GAT) by introducing an attention weighted neighborhood message aggregation mechanism.
Jiang et al.~\cite{gden} propose Graph Diffusion-Embedding Network (GDEN) by fusing feature diffusion and node embedding together for node representation and learning.

Although many spectral and spatial GCNs~\cite{GCN_suervy2} have been designed and demonstrated much more powerfully than traditional shallow graph representation models~\cite{piping2,piping3} on addressing various graph learning tasks, 
 the whole network's capacity and representation ability of existing GCNs are still limited from the deep learning perspective.
This is because deeper GCNs usually suffer from the over-smoothing issue due to their inherent graph signal propagation mechanisms~\cite{oversmooth1,oversmooth2}.
As we all know that as an important deep learning model, Convolution Neural Networks (CNNs) have achieved great success in many computer vision and machine learning tasks~\cite{imageclass2,imageseg3,speech1,sentence1}.
CNNs leverage deep multi-layer architecture with each convolution layer employing many learnable convolution filters (kernels)  to obtain rich feature representation, which thus can have high functional capacity to deal with the complex patterns in visual data analysis.
%
%
Some works have been proposed to build GCNs directly based on the classic CNNs~\cite{CNN_basedGCN1,CNN_basedGCN2,pscn}.
However, they generally lack of fully exploiting the graph structure information in its convolution definition and their flexibility and generality are usually limited~\cite{GCN_suervy3}.
Therefore, it is still natural to arise a question: \emph{Beyond graph signal processing, can we design graph convolutional network as flexibly as CNNs}?

%

With this problem in mind, we revisit previous GCNs and find that existing GCNs can be interpreted from the aspect of depthwise separable CNN~\cite{mobilenet} which is a remarkable CNN variant and has been widely used in ShuffleNet~\cite{shufflenet}, GhostNet~\cite{ghostnet},
ThunderNet~\cite{thundernet}, Involution~\cite{li2021involution} and also been employed to explain local-attention~\cite{han2021connection} and Transformers~\cite{Li2021}, etc. In this work, we aim to connect depthwise separable convolution with graph convolutions.
The core of depthwise separable convolution is to separate standard convolution operation into two parts, i.e., depthwise convolution and pointwise convolution. 
In this work, we show that previous popular GCNs (such as GCN~\cite{gcn} and GAT~\cite{gat}) indeed perform some specific depthwise separable convolution operations.
Specifically, we show that GCN~\cite{gcn} performs \emph{single {fixed}} depthwise convolution and \emph{multiple learnable} pointwise convolutions while GAT~\cite{gat} performs \emph{single parameterized} depthwise convolution and \emph{multiple learnable} pointwise convolutions.
 This explanation enables us to better understand the connections between GCNs and CNNs and further inspires us to design more powerful GCNs by leveraging the designing sprits of CNNs, such as Group Convolution~\cite{alexnet}, dilated convolution~\cite{dilatedconv} and Shufflenet~\cite{shufflenet} etc.
Based on our observation, in this work, we propose some new GCNs, named Unified GCNs (UGCNs).
As two showcases, we implement
two UGCNs, i.e., Separable UGCN (S-UGCN) and General UGCN (G-UGCN), for graph data representation and learning.
Promising experiments on several graph representation benchmarks demonstrate the effectiveness and some advantages of the proposed UGCNs.

The reminder of this paper is organized as follows.
In Section 2, we revisit the classici CNN and depthwise separable CNN.  
In Section 3, we present the connections between GCNs and depthwise separable CNNs. 
In Section 4, we introduce our Unified GCNs.  
We will propose the evaluation of our proposed UGCNs in Section 5.

\section{Reviewing CNNs}

In this section, we review the formulation of classic convolution and depthwise separable convolution in CNNs.


Let $\mathbf{X}=\{\mathbf{x}_{p}\}_{{p}\in\mathcal{P}}, \mathbf{x}_{p}\in\mathbb{R}^C$ denote the input features where $C$ is the feature dimension and $\mathcal{P}$ denotes the set of positions.
For instance,
in visual image data, $\mathcal{P}=\{(1,1),(1,2),\cdots (H,W)\}$ where $H$ and $W$ represent the height and width of input image respectively.
Let $\mathcal{N}({p})$ denotes the neighborhood position set of position ${p}$
and $\mathcal{K}({p},{p}',c), p'\in\mathcal{N}({p})$ denotes the convolution/filter weight.
Then, the basic convolution operation between $\mathbf{X}$ and $\mathcal{K}$ can be formulated as follows,
\begin{flalign}\label{EQ:C}
{z}_{{p}} = \sum^{C}_{c=1}\sum_{{p}'\in\mathcal{N}({p})}
\mathcal{K}({p},{p}',c)\cdot\mathbf{X}_{{p}',c}
\end{flalign}
where ${z}_{p}\in \mathbb{R}$ denotes the output feature value for position ${p}$.

\subsection{From Convolution to Convolutional Layer in CNNs}

In the past few years, Convolutional Neural Networks (CNNs)~\cite{lenet,googlenet,resnet} have achieved great success in machine learning and computer vision fields. 
The core point of CNNs is the design of convolutional layer (ConvL) for data feature extraction.
There are two main aspects to design ConvL in CNNs.

(1) \textbf{Weight Sharing:} The convolution weights $\mathcal{K}(:,:,:)$ in ConvL are \emph{shared} on all positions $\mathcal{P}$, i.e.,
\begin{equation}\label{EQ:11}
\forall  p, q\in \mathcal{P}: \mathcal{K}(p,:,:)=\mathcal{K}(q,:,:)
\end{equation}
For image data, the neighborhood $\mathcal{N}(p)$ of each position $p=(i,j)$ is  a grid.
 Therefore, one can use a fixed-size learnable kernel tensor $\mathbf{K}$ to
 conduct convolution operation.
 To be specific, for example, given an input feature map $\mathbf{X}\in\mathbb{R}^{H\times W\times C}$, where $\{H, W, C\}$  denote height, width and channel size respectively.
Then, the convolution operation in ConvL is formulated as follows,
\begin{flalign}\label{EQ:CL-kernel1}
\textbf{z}_{{p}}
=\sum^{C}_{c=1}\sum_{{p}'\in\mathcal{N}({p})}
 \mathbf{K}_{{p}'-{p}+\left\lfloor\frac{K}{2}\right\rfloor,c}\cdot\mathbf{X}_{{p}',c}
\end{flalign}
where
$\mathbf{K}\in\mathbb{R}^{K\times K\times C}$ denotes the learnable kernel tensor which is shared on all positions and $K$ is kernel size.

\begin{figure}[!htbp]
	\centering
\includegraphics[width=0.45\textwidth]{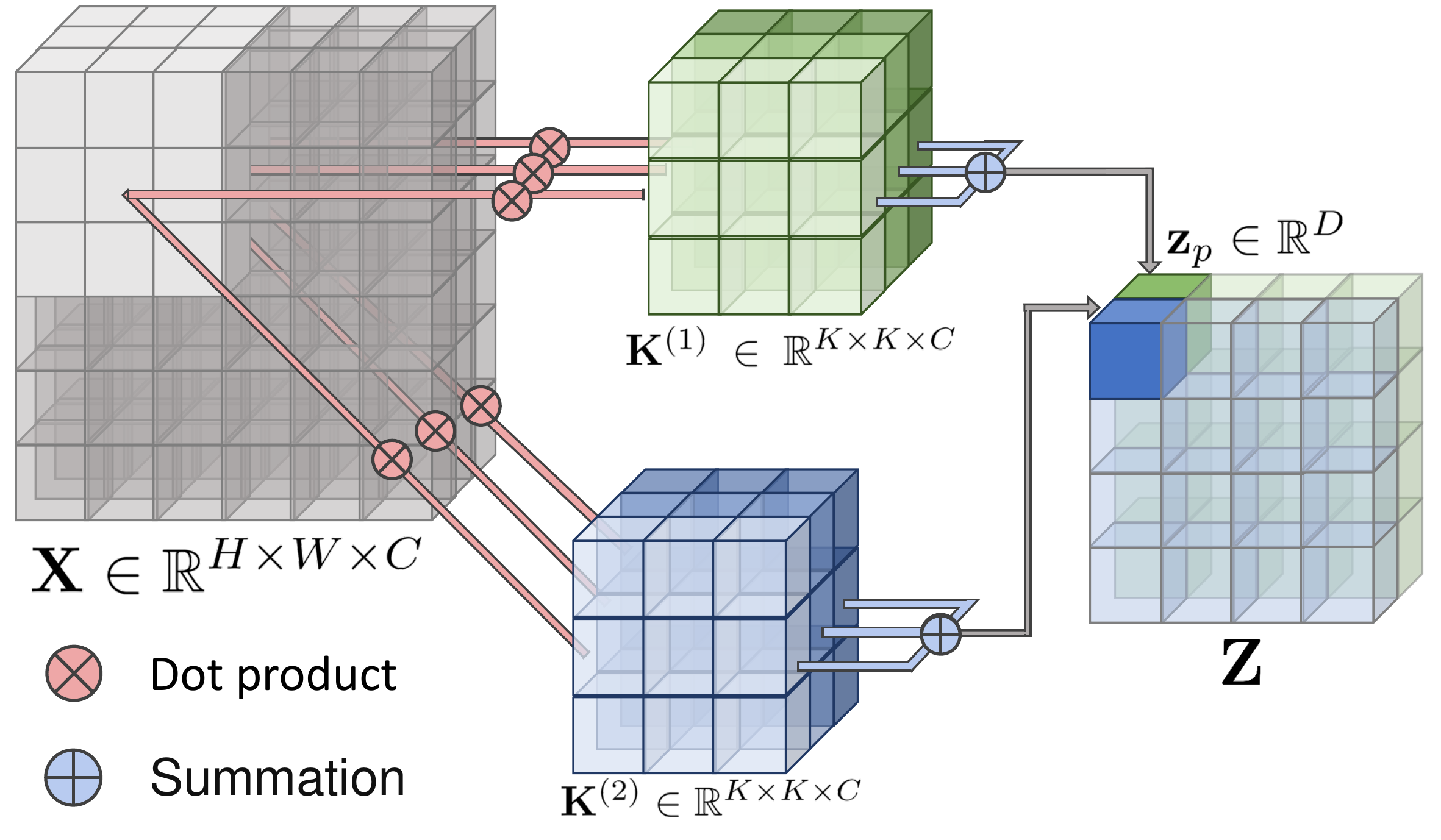}
	\caption{Illustration of a ConvL example ($H=6,W=6, K=3, C=3, D=2$) in CNNs.} 
\label{fig:CL}
\end{figure}

(2) \textbf{Multiple Filters:} ConvL further employs \emph{multiple} different kernel tensors $\big\{\mathbf{K}^{(1)}, \mathbf{K}^{(2)}\cdots \mathbf{K}^{(D)}\big\}$ to obtain rich feature representation, i.e.,
\begin{flalign}\label{EQ:CL-kernel2}
&\mathbf{Z}_{p,d}
=\sum^{C}_{c=1}\sum_{{p}'\in\mathcal{N}({p})}
 \mathbf{K}^{(d)}_{{p}'-{p}+\left\lfloor\frac{K}{2}\right\rfloor,c}\cdot\mathbf{X}_{{p}',c} \\
& for \,\,\ d=1, 2, \cdots D
\end{flalign}
%
where the output $\mathbf{Z}=\{\textbf{z}_p\}_{p\in \mathcal{P}}\in\mathbb{R}^{|\mathcal{P}|\times D}$ denotes the
extracted $D$-dimensional feature descriptors. 
All kernels $\big\{\mathbf{K}^{(1)}, \mathbf{K}^{(2)}, $ $\cdots, \mathbf{K}^{(D)}\big\}$ are learnable and can be optimized by minimizing the task-specific loss.
Figure~\ref{fig:CL} illustrates the implementation process of each ConvL in CNNs.

\textbf{Remark.}
In CNNs, to obtain nonlinear and compact feature representation,
the convolutional layer is usually followed by  pooling, activation and normalized layers  to
achieve subsampling, nonlinear enhancing and  feature normalization respectively. 
In this paper, we mainly focus on designing convolutional layers.
For concise presentation, we ignore these auxiliary layers (pooling, activation and normalized) in the following sections.

\subsection{Depthwise Separable Convolutional Layer}

One remarkable and efficient convolution variant is Depthwise Separable Convolution (DSConv) proposed in MobileNet~\cite{mobilenet}.
It is
commonly used in many lightweight CNNs, such as ShuffleNet~\cite{shufflenet}, Ghostnet~\cite{ghostnet},
ThunderNet~\cite{thundernet}, etc.
The core of DSConv layer is to separate classic ConvL into two parts, i.e., single Depthwise convolution (DConv) with shared spatial convolution kernel and multiple Pointwise convolutions (PConv) with different channel convolution kernels. 
The details of DConv and PConv are introduced as follows.

\subsubsection{DConv layer}

The aim of DConv~\cite{mobilenet} is to conduct the information mixing among different spatial positions.
The general formulation of DConv operation is follows,
\begin{flalign}\label{EQ:DCL}
\mathbf{Y}_{{p},c} =
\sum_{{p}'\in\mathcal{N}({p})}\mathcal{K}(p,{p}',c)\cdot\mathbf{X}_{{p}',c}
\end{flalign}
where $\mathbf{Y}\in\mathbb{R}^{|\mathcal{P}|\times C}$ denotes the output of DConv and $|\mathcal{P}|$ denotes number of all positions.
In DConv layer, the convolution weights are shared on all positions. Thus, similar to ConvL, it can be formulated as
\begin{flalign}\label{EQ:DCL-kernel}
\mathbf{Y}_{{p},c} =
\sum_{{p}'\in\mathcal{N}({p})}\mathbf{\hat{K}}_{{p}'-{p}+\left\lfloor\frac{K}{2}\right\rfloor,c}\cdot\mathbf{X}_{{p}',c}
\end{flalign}
where $\hat{\mathbf{K}}\in \mathbb{R}^{K\times K\times C}$ denotes the shared fixed-size kernel tensor.
Note that, in DConv layer, the convolution operation is only conducted on spatial dimension. This is different from standard convolution operation in which the final output is obtained by summating all channels, as shown in Eqs.(\ref{EQ:C},\ref{EQ:CL-kernel1}).


\subsubsection{PConv layer}

After the above DConv, 
pointwise convolution (PConv)~\cite{mobilenet} is further adopted in depthwise separable convolution for feature extraction.
The aim of PConv is to conduct  information mixing among different channels.
The general formulation of PConv operation is follows,
\begin{flalign}\label{EQ:PCL}
\mathbf{z}_{{p}} =
\sum^{C}_{c=1}\kappa(c)\cdot\mathbf{Y}_{p,c}
\end{flalign}
where $\mathbf{Y}$ is the output of the above DConv (Eq.(\ref{EQ:DCL})).
Note that, PConv does not aggregate the information from the neighbors $\mathcal{N}({p})$ of each position $p$. 

\begin{figure}[!htbp]
	\centering
\includegraphics[width=0.47\textwidth]{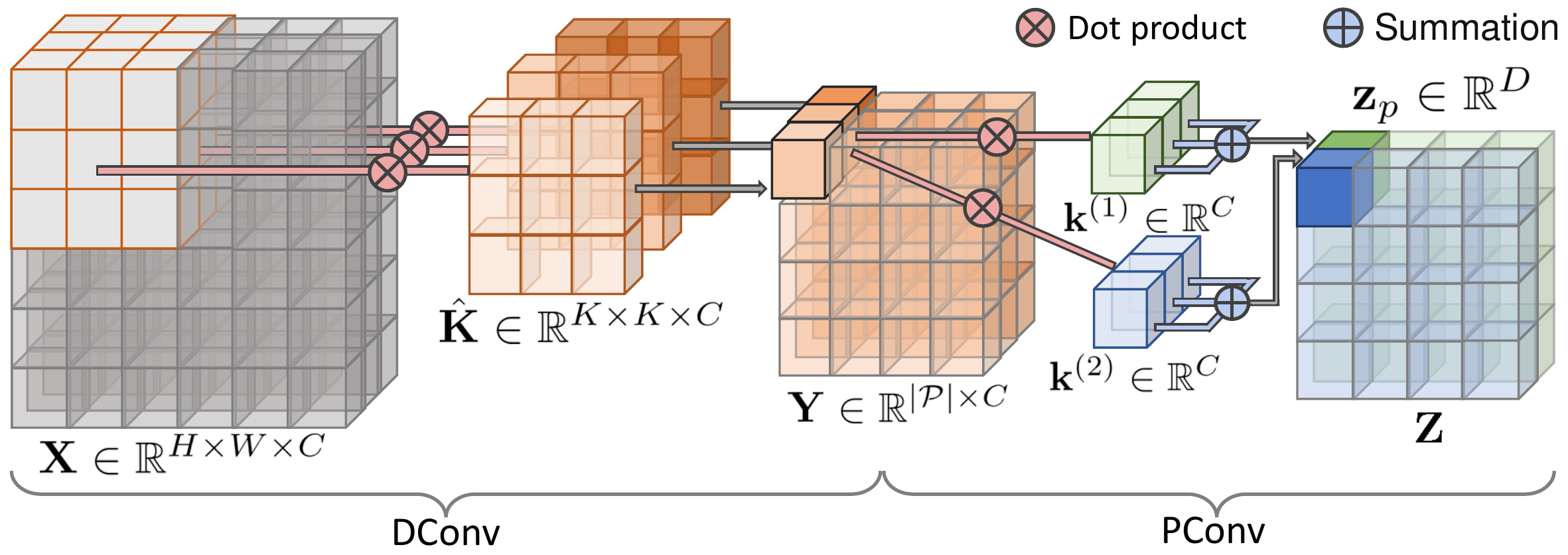}
	\caption{Illustration of a DSConv example ($H=6, W=6, C=3, D=2, K=3$) in depthwise separable CNNs.}\label{fig:mobilenet}
\end{figure}

Also, to obtain rich feature representation, PConv layer employs \emph{multiple} different kernel weights $\kappa^{(d)}(c), d=1, 2\cdots D$ and each of them is shared on all positions, i.e., 
\begin{flalign}\label{EQ:PCL-kernel1}
\mathbf{Z}_{p,d}
=\sum^{C}_{c=1}\kappa^{(d)}(c)\cdot\mathbf{Y}_{{p},c},\,\, for \,\,\, d = 1, 2\cdots D
\end{flalign}
For instance, in image representation, one can use \textbf{multiple} learnable kernel matrix representations $\mathbf{k}^{(d)} \in\mathbb{R}^{C}, d = 1, 2\cdots D$ and Eq.(\ref{EQ:PCL-kernel1}) is reformulated as
\begin{flalign}\label{EQ:PCL-kernel2}
\mathbf{Z}_{p,d}
=\sum^{C}_{c=1}\mathbf{k}^{(d)}_{c}\cdot\mathbf{Y}_{{p},c},\,\ for \,\,\, d = 1, 2\cdots D
\end{flalign}
%
Figure~\ref{fig:mobilenet} demonstrates the entire operation process of DSConv~\cite{mobilenet} which contains both DConv and PConv layers.


\section{Connecting GCNs with DSConv}

Graph convolutional networks (GCNs) have been widely used in various graph data representation and learning tasks, such as,
computer vision~\cite{MGCNCO,wangxixi}, semi-supervised node classification~\cite{gcn,gat} and graph classification~\cite{gin,diffpool}.
In this section, we show that existing GCNs generally conduct DSConv with some specific constraints.
We take two commonly used GCNs (GCN~\cite{gcn} and GAT~\cite{gat}) as showcases. Some other GCNs can be similarly explained.

\subsection{Explaining GCN with DSConv}

Kipf et al.~\cite{gcn} propose a widely used Graph Convolutional Network (GCN).
Below, we first review the general formulation of graph convolutional (GC) layer in GCN~\cite{gcn}. Then, we
analyze the connection between GC Layer and DSConv (Eqs.(\ref{EQ:DCL},\ref{EQ:PCL})).

\subsubsection{ Graph Convolution (GC) layer}

Given an input graph data $\mathcal{G}(\mathbf{X},\mathbf{A})$ where
$\mathbf{X}=\{\mathbf{x}_{p}\}_{{p}\in\mathcal{P}}, \mathbf{x}_{p}\in\mathbb{R}^C$ denotes the collection of node features and
 $\mathcal{P}$ denotes the position (node) set\footnote{For graph data, each position denotes each node of the graph.}
  and thus $\mathbf{X}\in\mathbb{R}^{N\times C}$ where $N=|\mathcal{P}|$.
 $\mathbf{A}\in\mathbb{R}^{N\times N}$ denotes the adjacency matrix.
Kipf et al.~\cite{gcn} propose to define a popular graph convolution operation as,
\begin{flalign}\label{EQ:GCL}
\mathbf{Z} = \mathbf{D}^{-\frac{1}{2}}\tilde{\mathbf{A}}\mathbf{D}^{-\frac{1}{2}}\mathbf{X}\mathbf{W}
\end{flalign}
where $\mathbf{Z}=\{\mathbf{z}_p\}_{p\in \mathcal{P}}\in\mathbb{R}^{N\times D}$ denotes the output node features.
Here $\tilde{\mathbf{A}} = \mathbf{A}+\mathbf{I}$ and $\mathbf{D}_{p,p}=\sum_{p'\in\mathcal{N}(p)}\tilde{\mathbf{A}}_{p,p'}$. 
$\mathbf{W}\in\mathbb{R}^{C\times D}$ denotes the learnable weight matrix.
Let $\mathbf{\hat{A}} =  \mathbf{D}^{-\frac{1}{2}}\tilde{\mathbf{A}}\mathbf{D}^{-\frac{1}{2}}$. Then, Eq.(\ref{EQ:GCL}) can be rewritten as
\begin{flalign}\label{EQ:GCL-e}
\mathbf{Z}_{{p},d}= \sum_{c=1}^{C}\sum_{{p}'\in\mathcal{N}({p})}\mathbf{\hat{A}}_{{p},{p}'}\cdot\mathbf{X}_{{p}',c}\cdot\mathbf{W}_{c,d}
\end{flalign}
where $d = 1, 2\cdots D$.

\subsubsection{Connection between GC and DSConv}

\begin{figure}[!htbp]
	\centering
\includegraphics[width=0.47\textwidth]{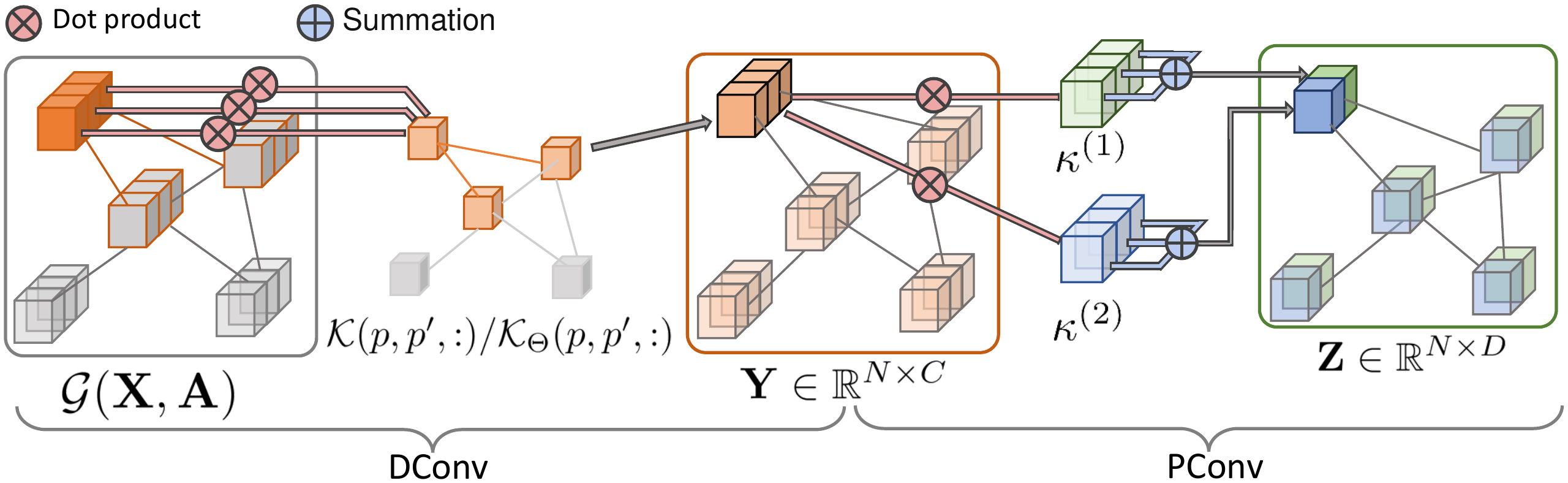}
	\caption{Illustration of a GC/GAT layer example ($N=5$, $C=3$, $D=2$) in GCN/GAT.}\label{fig:GCL}
\end{figure}

We show that the above GC operation Eqs.(\ref{EQ:GCL},\ref{EQ:GCL-e})  can be regarded as a kind of DSConv in \S 2.2, i.e., it indeed conducts single DConv operation (Eq.(\ref{EQ:DCL})) and \textbf{multiple} PConv operations (Eq.(\ref{EQ:PCL})).
To be specific, the above GC (Eq.(\ref{EQ:GCL-e})) can be decomposed into two steps, i.e.,
\begin{align}\label{EQ:GCL-2}
&\mathbf{DConv}:\,\,\, \mathbf{Y}_{{p},c} =
\sum_{{p}'\in\mathcal{N}({p})}\mathcal{K}(p,{p}',c)\cdot\mathbf{X}_{{p}',c}\\
&\mathbf{PConv}:\,\,\, \mathbf{Z}_{{p},d} =
\sum^{C}_{c=1}\kappa^{(d)}(c)\cdot\mathbf{Y}_{p,c} \,\, for \,\, d = 1 \cdots D
\end{align}
where $\mathcal{K}(p,p',c)$ and
$\kappa^{(d)}(c)$
in Eqs.(13,14) are defined as
\begin{flalign}\label{EQ:define-gcl}
&\mathcal{K}(p,p',:)=\Big[\mathbf{\hat{A}}_{p,p'}, \mathbf{\hat{A}}_{p,p'},\cdots,\mathbf{\hat{A}}_{p,p'}\Big]\\
&\kappa^{(d)}(c)= \mathbf{W}_{c,d}
\end{flalign}
Therefore, we can observe that GC can be regarded as DSConv with two specific constraints:
\begin{itemize}
  \item Instead of learning convolution weights, GC adopts the \textbf{fixed} convolution weights, i.e., $\mathcal{K}(p,p',c)=\mathbf{\hat{A}}_{p,p'}, p'\in\mathcal{N}(p)$ in its convolution operation.
  \item GC \textbf{shares} the depth-wise convolution weights on all channel dimensions, i.e., $\mathcal{K}(p,p',c)=\mathbf{\hat{A}}_{p,p'}$ for $ c=1, 2 \cdots C$,
   as similar as that in involution operation~\cite{li2021involution}. 
\end{itemize}
Figure~\ref{fig:GCL} demonstrates the whole implementation process of GC layer from DSConv perspective which involves both DConv and multiple PConv operations.

\subsection{Explaining GAT with DSConv}

Veli{\v{c}}kovi{\'c} et al.~\cite{gat} propose a more flexible Graph Attention Network (GAT) for graph data representation and learning.
Below, we first review the general formulation of graph attention (GAT) layer and then analyze the connection between GAT and DSConv.  

\subsubsection{Graph Attention (GAT) Layer}

Given an input graph data $\mathcal{G}(\mathbf{X},\mathbf{A})$ where
$\mathbf{X}=\{\mathbf{x}_{p}\}_{{p}\in\mathcal{P}}, \mathbf{x}_{p}\in\mathbb{R}^C$ denotes the collection of node features and
 $\mathcal{P}$ denotes the position (node) set and thus $\mathbf{X}\in\mathbb{R}^{N\times C}$.
 $\mathbf{A}\in\mathbb{R}^{N\times N}$ denotes the adjacency matrix.
%
GAT~\cite{gat} first learns an edge attention $\alpha_{p,p'}$ for each edge  as follows,
\begin{flalign}\label{EQ:define-alpha-gat}
\alpha_{p,p'}=\left\{
\begin{aligned}
&f_{\Theta}(\mathbf{x}_{p},\mathbf{x}_{p'}) &\mathrm{\,\,\,if}\ \ p'\in\mathcal{N}(p)\\
&\,\,\,\,\,\,\,\,\,0 &\mathrm{otherwise}
\end{aligned}
\right.
\end{flalign}
where $\Theta$ denotes the learnable parameters and $\mathcal{N}(p)$ denotes the neighborhood set of position (node) $p$.
Then, GAT uses the learnable $\alpha$ to replace the fixed adjacency matrix $\mathbf{A}$ to achieve the layer-wise message propagation as
\begin{flalign}\label{EQ:GATL}
\mathbf{Z}= \alpha\mathbf{X}\mathbf{W}
\end{flalign}
where $\mathbf{Z}=\{\mathbf{z}_p\}_{p\in\mathcal{P}}\in\mathbb{R}^{N\times D}$ denotes the output features and $\mathbf{W}\in\mathbb{R}^{C\times D}$ denotes the learnable weight matrix.
The above Eq.(\ref{EQ:GATL}) can be rewritten as
\begin{flalign}\label{EQ:GATL-e}
\mathbf{Z}_{p,d}= \sum_{c=1}^{C}\sum_{p'\in\mathcal{N}(p)}\alpha_{p,p'}\cdot\mathbf{X}_{p',c}\cdot\mathbf{W}_{c,d}
\end{flalign}
where $d = 1, 2\cdots D$.
\subsubsection{Connection between GAT and DSConv}
We show that the above GAT can be regarded as a kind of DSConv.
Specifically, GAT Eq.(\ref{EQ:GATL-e}) can be decomposed into DConv and PConv steps, i.e.,
\begin{flalign}\label{EQ:GATL-2}
&\mathbf{DConv:}\,\,\, \mathbf{Y}_{{p},c} =
\sum_{p'\in\mathcal{N}(p)}\mathcal{K}_{\Theta}(p,{p}',c)\cdot\mathbf{X}_{{p}',c}\\
&\mathbf{PConv:}\,\,\, \mathbf{Z}_{{p},d} =
\sum^{C}_{c=1}\kappa^{(d)}(c)\cdot\mathbf{Y}_{p,c},\,for \,\,\, d = 1 \cdots D
\end{flalign}
where $\mathcal{K}_{\Theta}(p,p',c)$  and $\kappa^{(d)}(c)$ in Eqs.(20, 21) are respectively defined as
\begin{flalign}\label{EQ:define-gat}
&\mathcal{K}_{\Theta}(p,p',:)=\Big[\alpha_{p,p'},\alpha_{p,p'},\cdots,\alpha_{p,p'}\Big]\\
&\kappa^{(d)}(c) = \mathbf{W}_{c,d}
\end{flalign}
Therefore, we can observe that  GAT can be regarded as a kind of DSConv with three specific constraints:
\begin{itemize}
  \item Instead of learning depth-wise convolution weights $\mathcal{K}(p,p',c)$ completely in DConvL, GAT learns depth-wise convolution weights $\mathcal{K}_{\Theta}(p,p',c)$ \textbf{parameterized} by $\Theta$, i.e., GAT learns convolution weights $\mathcal{K}_{\Theta}(p,p',c)$ indirectly by learning the parameter $\Theta$.  
      \item Instead of sharing convolution weights $\mathcal{K}(p,p',c)$ on all positions directly in CNNs, GAT \textbf{shares} the learnable parameters $\Theta$ in $\mathcal{K}_{\Theta}(p,p',c)$ on all positions.
  \item  GAT also \textbf{shares} the depth-wise convolution weights on all channels, i.e., $\mathcal{K}_{\Theta}(p,p',c)=\alpha_{p,p'}=f_{\Theta}(\mathbf{x}_{p},\mathbf{x}_{p'})$, for all $c\in \{1, 2 \cdots C\}$, as similar as that in self-attention~\cite{han2021connection} and involution operation~\cite{li2021involution}. 
\end{itemize}
Figure~\ref{fig:GCL} demonstrates the whole implementation process of GAT layer from DSConv perspective which involves both DConv and multiple PConv operations.


\section{Our Proposed Unified GCNs}

Based on the above observations, we can then develop some new GCNs, named Unified GCNs (UGCNs)  for graph data representation by borrowing the designing sprits from CNNs, such as Group Convolution~\cite{alexnet}, ShuffleNet~\cite{shufflenet} and GoogleNet~\cite{googlenet}.
As two showcases, in this paper, we implement two UGCNs, i.e., Separable UGCN (S-UGCN) and General UGCN (G-UGCN). 


\subsection{Separable UGC Layer}

As discussed in Eqs.(15,22), both GCN and GAT shares the depth-wise convolution weights on all channels.
It is known that in classic CNNs and depthwise separable CNN (MobileNet)~\cite{mobilenet}, each channel conducts specific convolution operation to obtain rich features. 
Inspired by this,
similar to MobileNet~\cite{mobilenet}, we propose our Separable UGCN (S-UGCN) which defines a specific convolution filter for each channel.

\begin{figure}[!htbp]
	\centering
\includegraphics[width=0.47\textwidth]{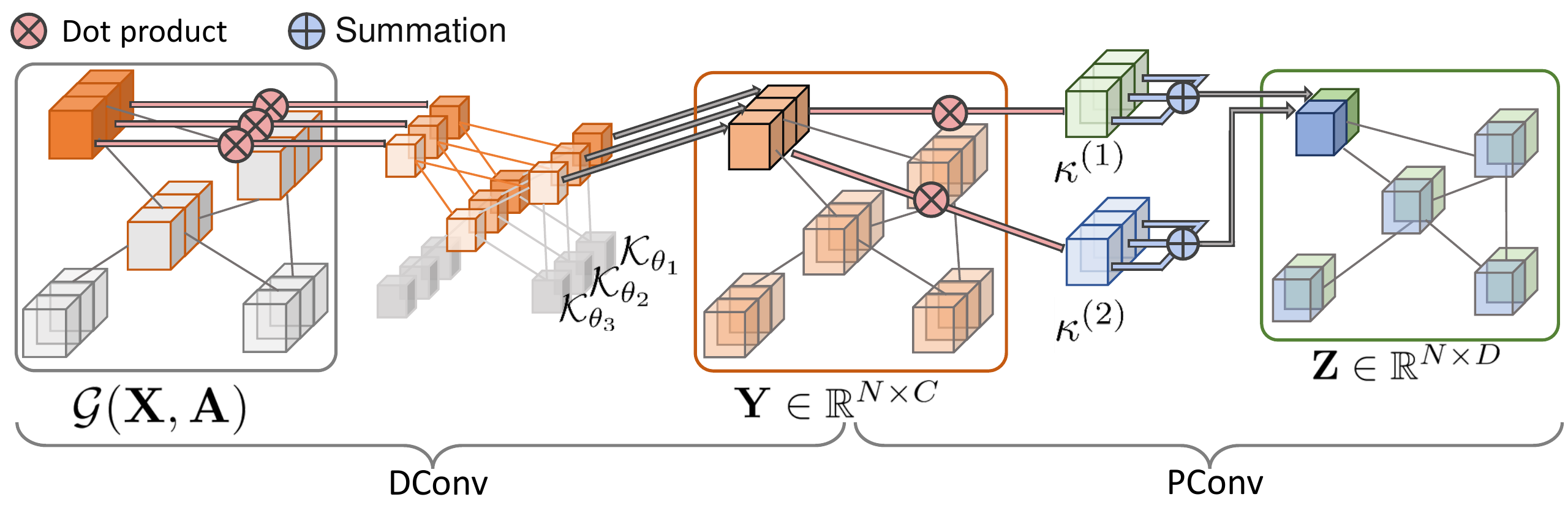}
	\caption{Illustration of a S-UGC example ($N=5, C=3, D=2$) in S-UGCN.}\label{fig:sgcl}
\end{figure}

Specifically, each S-UGC layer in S-UGCN involves both depthwise and pointwise convolution operations.
Given an input graph data $\mathcal{G}(\mathbf{X},\mathbf{A})$ where $\mathbf{X}\in\mathbb{R}^{N\times C}$ represents the node feature matrix and
$\mathbf{A}\in\mathbb{R}^{N\times N}$ denotes the adjacency matrix,
our S-UGC layer involves both depthwise and pointwise convolution operations which are generally defined  as follows,
\begin{flalign}\label{EQ:DSGCL-2}
&\mathbf{DConv:}\,\,\, \mathbf{Y}_{{p},c} =
\sum_{{p}'\in\mathcal{N}({p})}\mathcal{K}_{\theta_c}(p,p',c)\cdot\mathbf{X}_{{p}',c}\\
&\mathbf{PConv:}\,\,\, \mathbf{Z}_{{p},d} =
\sum^{C}_{c=1}\kappa^{(d)}(c)\cdot\mathbf{Y}_{p,c},\, for \,\,\, d = 1 \cdots D
\end{flalign}
where $\mathbf{Z}\in\mathbb{R}^{N\times D}$ denotes the outputs of our S-UGC layer.
$\mathcal{K}_{\theta_{c}}(p,p',c)$ denotes the specific convolution filter for the $c$-th channel/dimension and $\mathrm{\theta}_{c}$ is the learnable parameter which is shared on all positions $\mathcal{P}$. 
In practical, $\mathcal{K}_{\theta_{c}}(p,p',c)$ and $\kappa(c)$ in Eqs.(24, 25) can be defined as
\begin{flalign}\label{EQ:define-dsgcl}
\mathcal{K}_{\theta_{c}}(p,p',c)&=f_{\theta_{c}}(\mathbf{X}_{p,c},\mathbf{X}_{p',c})\\
\kappa^{(d)}(c) &= \mathbf{W}_{c,d}
\end{flalign}
where $\mathbf{W}\in \mathbb{R}^{C\times D}$ is a weight matrix.
Let
\begin{flalign}\label{EQ:define-alpha}
\alpha_{p,p',c}=\left\{
\begin{aligned}
&f_{\theta_{c}}(\mathbf{X}_{p,c},\mathbf{X}_{p',c}) &\mathrm{\,\,\,if}\,\ p'\in\mathcal{N}(p)\\
&\,\,\,\,\,\,\,\,\,0 &\mathrm{otherwise}
\end{aligned}
\right.
\end{flalign}
Then, our S-UGC can be formulated compactly as
%
\begin{flalign}\label{EQ:DSGCL}
&\mathbf{Z}_{p,d} = \sum_{c=1}^{C}\sum_{p'\in\mathcal{N}(p)}\alpha_{p,p',c}\cdot\mathbf{X}_{p',c}\cdot\mathbf{W}_{c,d}\\
&for \,\,\, d = 1, 2\cdots D \nonumber
\end{flalign}
%
Figure~\ref{fig:sgcl} demonstrates the whole implementation process of the proposed S-UGC layer.
Note that, comparing with GAT (Eq.(\ref{EQ:GATL-e})), S-UGC employs specific filter for each channel and thus can obtain richer features to represent complex pattern in graph data representation and learning.

\subsection{General UGC Layer}

\begin{figure}[!htbp]
	\centering
\includegraphics[width=0.45\textwidth]{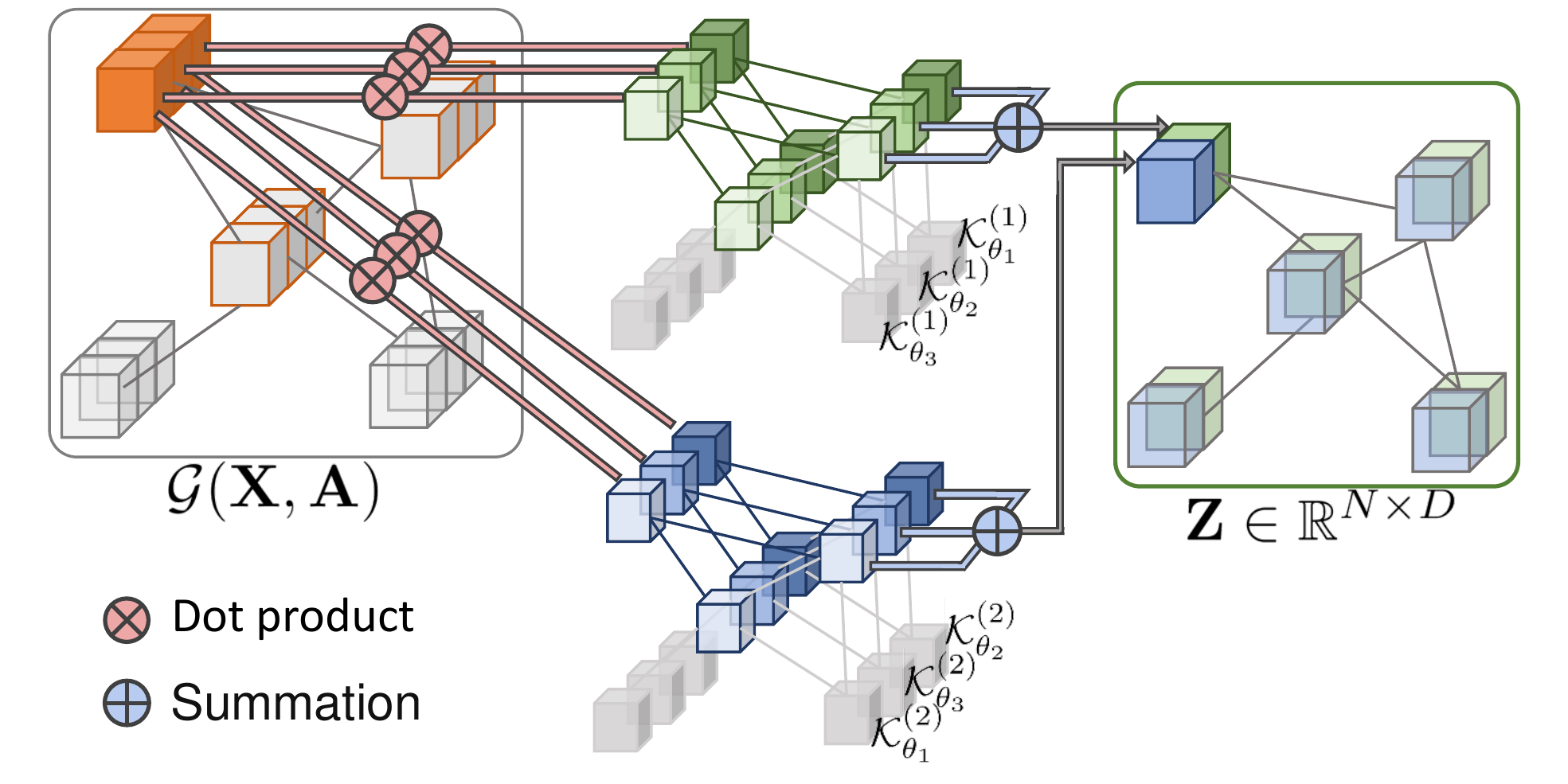}
	\caption{Illustration of a G-UGC example ($N=5$, $C=3$, $D=2$) in G-UGCN.}\label{fig:non-sgcl}
\end{figure}

As discussed, both GCN and GAT can be connected with DSConv. The above proposed S-UGCN is also derived from DSConv.
Therefore, it is natural to arise a question: beyond DSConv, can we design graph convolutional layer similarly to classic CNNs directly?
Here, we develop a simpler and more general network, named General UGCN (G-UGCN).
The key aspect of G-UGCN is that it employs multiple general convolutions directly to obtain rich features.
Formally, the proposed G-UGC layer is formulated as
%
%
\begin{flalign}\label{EQ:UGCL}
&\mathbf{Z}_{p,d} =
\sum_{c=1}^{C}\sum_{{p}'\in\mathcal{N}({p})}\mathcal{K}^{(d)}_{\theta_{c}}(p,{p}',c)\cdot\mathbf{X}_{{p}',c}\\
&for \,\,\, d = 1, 2\cdots D \nonumber
\end{flalign}
where $\mathbf{Z}\in\mathbf{R}^{N\times D}$ denotes the output features.
$\mathcal{K}^{(d)}_{\theta_{c}}(p,p',c)$ denotes the $d$-th convolution filter and $\mathrm{\theta}_{c}$ is the learnable parameter which is shared on all positions in each layer.
In practical, $\mathcal{K}^{(d)}_{\theta_{c}}(p,p',c)$ can be defined as
\begin{flalign}\label{EQ:define-gugcl}
\mathcal{K}^{(d)}_{\theta_{c}}(p,p',c)&=f_{\theta_c^{(d)}}(\mathbf{X}_{p,c},\mathbf{X}_{p',c})
\end{flalign}
%
Moreover, let
\begin{flalign}\label{EQ:define-alpha}
\alpha^{(d)}_{p,p',c}=\left\{
\begin{aligned}
&f_{\theta_c^{(d)}}(\mathbf{X}_{p,c},\mathbf{X}_{p',c}) &\mathrm{\,\,\,if}\,\ p'\in\mathcal{N}(p)\\
&\,\,\,\,\,\,\,\,\,0 &\mathrm{otherwise}
\end{aligned}
\right.
\end{flalign}
%
Then, we can rewritten Eq.(\ref{EQ:UGCL}) as
\begin{flalign}\label{EQ:UGCL-kernel}
&\mathbf{Z}_{p,d} = \sum_{c=1}^{C}\sum_{p'\in\mathcal{N}(p)}\alpha^{(d)}_{p,p',c}\cdot\mathbf{X}_{p',c}\\
&for \,\,\, d = 1, 2\cdots D \nonumber
\end{flalign}
Note that,  the weight matrix $\mathbf{W}$ which is commonly used in classic GCs is not required in our G-UGC definition.
Figure~\ref{fig:non-sgcl} demonstrates the whole operation process of G-UGC layer.

\subsection{Network Architecture and Implement Detail}

In this section, we demonstrate the implement detail and our network architecture.
The convolution weights in Eqs.(\ref{EQ:define-dsgcl},\ref{EQ:define-gugcl}) can be defined in many ways.
In this paper, 
we propose to simply define the convolution weight $\mathcal{K}_{\theta_{c}}(p,p',c)$ in Eq.(\ref{EQ:define-dsgcl}) as
\begin{flalign}\label{EQ:define-alpha-sugc}
f_{\theta_c}(\mathbf{X}_{p,c},\mathbf{X}_{p',c})= \mathrm{Softmax}\Big(\sigma\big({\theta_c}^{T}[\mathbf{X}_{p,c}\|\mathbf{X}_{p',c}]\big)\Big)
\end{flalign}
where $\|$ denotes the concatenation operation and $\theta_c\in \mathbb{R}^{2\times 1}$ denotes the
parameter vector.
$\sigma(\cdot)$ denotes the activation function and we use LeakReLU in our experiments.
Softmax activation function is used to normalize the outputs.
Similarly, the convolution weight $\mathcal{K}^{(d)}_{\theta_{c}}(p,p',c)$ in Eq.(\ref{EQ:UGCL}) is defined as
\begin{flalign}\label{EQ:define-alpha-gugc}
f_{\theta_c^{(d)}}(\mathbf{X}_{p,c},\mathbf{X}_{p',c})=\mathrm{Softmax}\Big(\sigma\big({\theta_c^{(d)}}^{T}[\mathbf{X}_{p,c}\|\mathbf{X}_{p',c}]\big)\Big)
\end{flalign}
where $\theta^{(d)}_c\in \mathbb{R}^{2\times 1}, d=1,2\cdots D$ denotes the $D$ parameter vectors.
Note that, due to the ability of BatchNorm (BN) layer to achieve normalization
on the graph classification task, we can remove Softmax and activation functions to obtain a simper output on graph classification task.

\begin{figure}[!htbp]
	\centering
\includegraphics[width=0.4\textwidth]{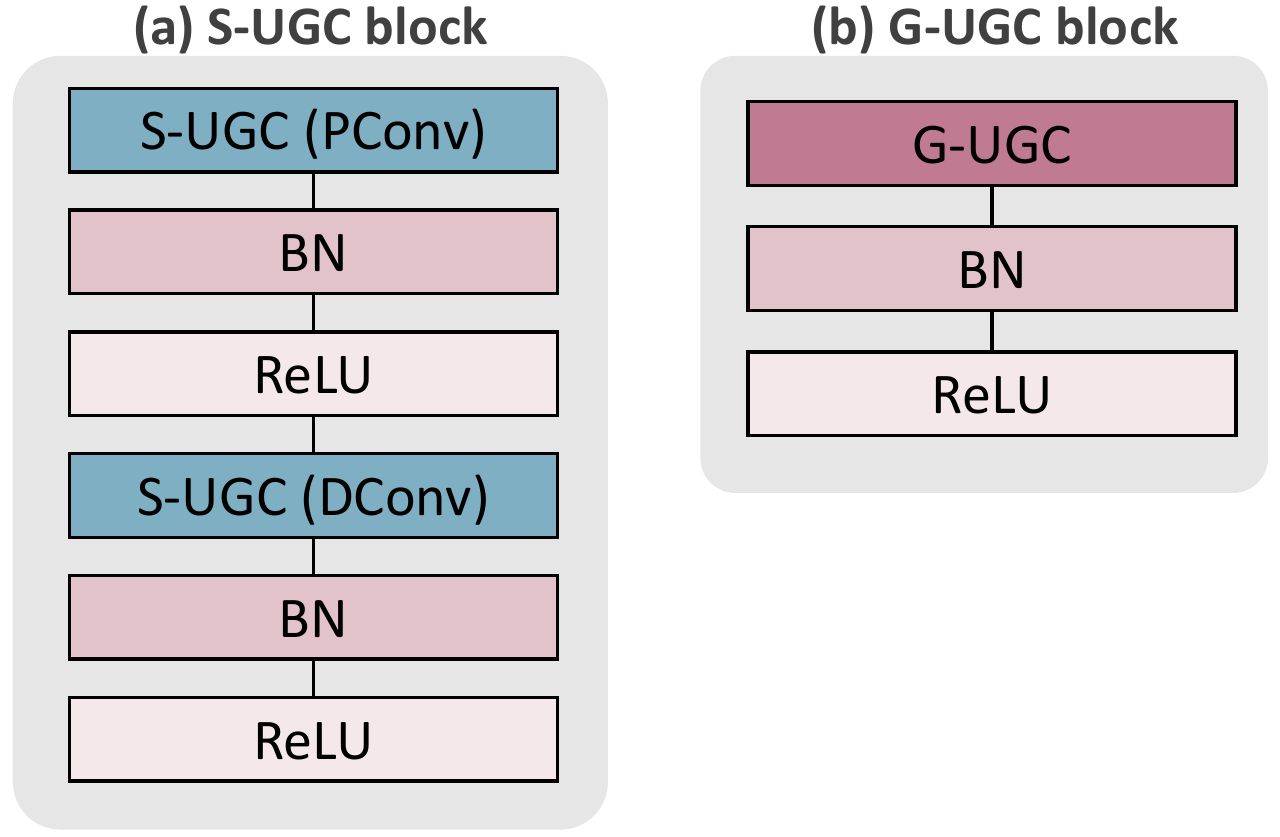}
	\caption{Left: S-UGC block involves DConv and PConv followed by BatchNorm (BN) and ReLU.
             Right: G-UGC block involves G-UGC followed by BN and ReLU.}\label{fig:block}
\end{figure}

\begin{figure*}[!htbp]
	\centering
\includegraphics[width=0.9\textwidth]{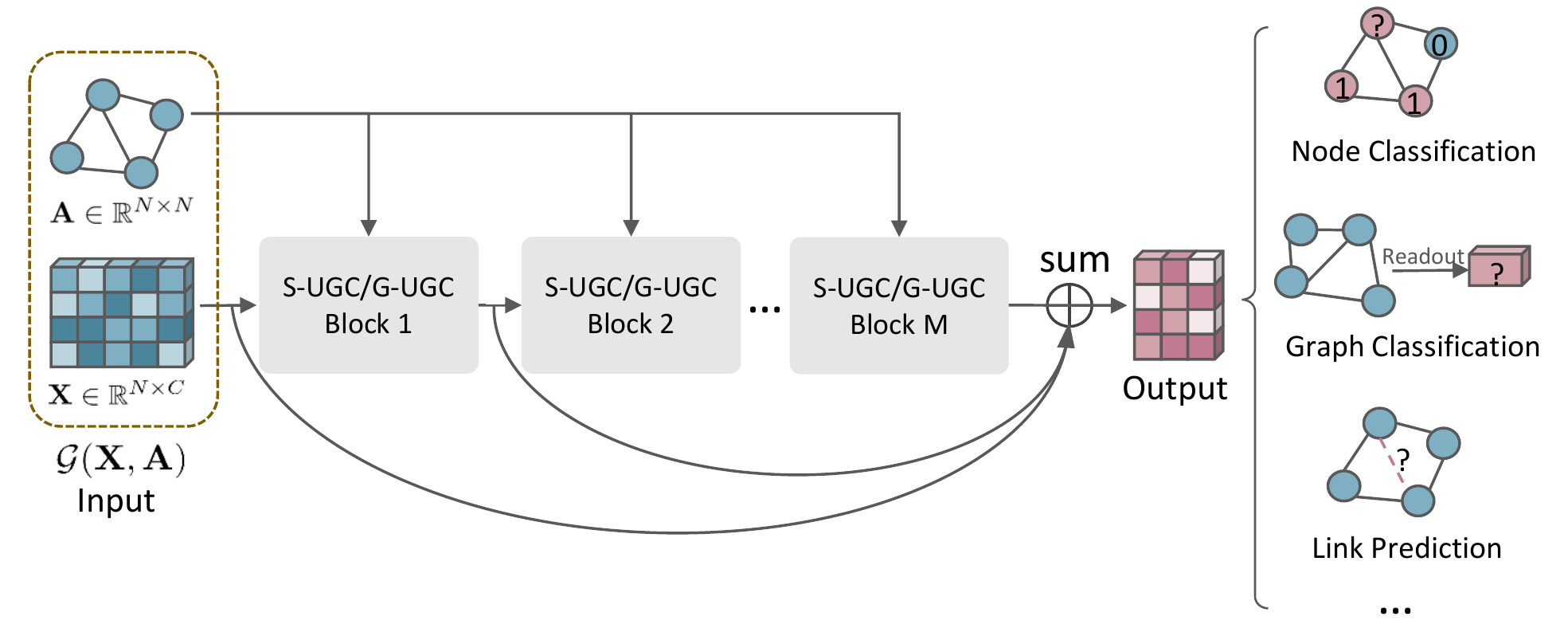}
	\caption{The network architecture of S-UGCN and G-UGCN.}\label{fig:architecture}
\end{figure*}

Based on the proposed UGCs (S-UGC, G-UGC), we can design UGC networks as flexibly as CNNs.
In this paper, as a showcase, we propose a kind of Unified GCN (UGCN) for graph data learning tasks.
Figure \ref{fig:architecture} shows the whole architecture of the proposed UGCN.
It contains multiple blocks in which each block (S-UGC and G-UGC block) is built as follows.
As illustrated in Figure \ref{fig:block} (a), S-UGC block involves DConv and PConv layers which are followed by the normalization and activation layer respectively, as suggested in CNNs~\cite{mobilenet,shufflenet}.
As shown in Figure \ref{fig:block} (b), G-UGC block contains G-UGC layer followed by the normalization and activation layers.
In our experiments, we use BatchNorm (BN) and ReLU as the normalization and activation layers respectively. 
Besides, as shown in Figure \ref{fig:architecture}, our UGCNs (S-UGCN and G-UGCN) also employ the skip connection strategy which summarizes all the outputs from different layers to fuse the high-level and low-level feature information, as commonly used in classic CNNs~\cite{densenet,resnet,unet}.
The proposed S-UGCN and G-UGCN can be applied on many graph representation and learning tasks, such as node classification, graph classification, link prediction, etc.
All the convolution parameters in S-UGCN and G-UGCN are trained by using Adam~\cite{adam} algorithm to optimize the specific loss function.

\section{Experiment}

\subsection{Experimental Setting}
\subsubsection{Datasets}

We utilize six graph classification datasets~\cite{datasets}, i.e., three bioinformatics datasets including MUTAG, PTC-MR and PROTEINS and three social network datasets including IMDB-BINARY, IMDB-MULTI and COLLAB.
The introduction of these datasets are demonstrated in Table \ref{tab:dataset}.
Following pervious work~\cite{gin},
we perform 10-fold cross-validation~\cite{chang2011libsvm} and report the average and standard deviation of accuracies. 

\subsubsection{Comparison methods}

To evaluate the effectiveness and benefit of the proposed UGCNs,
we first compare the proposed S-UGCN and G-UGCN with many state-of-the-art graph neural networks (GNNs), i.e.,
Graph Convolutional Network (GCN)~\cite{gcn}, Graph Attention Networks (GAT)~\cite{gat},
Deep Graph Convolutional Neural Network (DGCNN)~\cite{dgcnn}, PATCHY-SAN (PSCN)~\cite{pscn}, Graph Capsule Convolutional Neural Networks (GCCNN)~\cite{gccnn}, Graph Isomorphism Network (GIN)~\cite{gin},  Backtrackless Aligned-Spatial Graph
Convolutional Networks (BASGCN)~\cite{basgcn}, Dual Attention Graph Convolutional
Networks (DAGCN)~\cite{dagcn}, DropGIN~\cite{dropgnn}, Factorizable Graph Convolutional Networks (FactorGCN)~\cite{factorgcn} and Wasserstein Embedding For Graph Learning (WEGL)~\cite{wegl}.
Also, we compare our approaches with some other recent graph representation methods which contain four contrastive learning frameworks, i.e., Contrastive multi-view graph representation learning framework (CMRL)~\cite{cmrl}, InfoGraph~\cite{infograph}, GraphCL~\cite{graphcl} and Curriculum Contrastive learning framework (CUCO)~\cite{cuco}, and two graph pooling methods, i.e., Differentiable Pooling (DiffPool)~\cite{diffpool} and Graph Multiset Transformer (GMT)~\cite{gmt}.
In addition, we compare our methods with some traditional graph kernel based methods, i.e.,
Jensen-Tsallis q-difference kernel (JTQK)~\cite{jtqk}, Weisfeiler-Lehman (WL-Subtree)~\cite{wlsubtree},
Shortest Path kernel based on Core variants (CORE SP)~\cite{coresp} and Graphlet count Kernel (GK)~\cite{gk}.
For GCN~\cite{gcn} and GAT~\cite{gat},
we use the same  parameter setting with our proposed UGCNs.
Except GCN~\cite{gcn} and GAT~\cite{gat}, the performance of all other comparison results have been reported in their published papers and here we use them directly.

\subsubsection{Parameter settings}

Similar to pervious graph classification works~\cite{gin,gatgcc}, we use the sum readout function to obtain the final graph embedding.
The learning rate is set to $0.001$. The hidden size is selected from $\{16,32,64\}$ and the batch size is selected from $\{32,128\}$ to obtain better results.
The dropout ratio after the dense layer is selected from $\{0.0,0.5\}$.
The number of blocks in S-UGCN and G-UGCN are selected from $\{5,10\}$ to get better performance.
Besides, to stabilize the learning process, S-UGCN also employs a multi-head strategy similar to GAT~\cite{gat}.
The head number is selected from $\{1,4,8\}$.
Following pervious works~\cite{gin,dropgnn}, we fix the maximum epoch to $500$ and report the averaged results across the ten folds with cross-validation.
\begin{table*}[!ht]
    \centering
    \caption{The information of graph datasets.}
    \label{tab:dataset}
    \begin{tabular}{c|c|c|c|c|c|c|c}
    \hline
    \hline
      \multicolumn{2}{c|}{\multirow{2}{*}{Datasets}}&\multicolumn{3}{c|}{Bioinformatics datasets}&\multicolumn{3}{c}{Social datasets}\\
  \cline{3-8}
  \multicolumn{2}{c|}{} & MUTAG  & PTC-MR & PROTEINS & IMDB-BINARY & IMDB-MULTI & COLLAB\\
    \hline
   \multicolumn{2}{c|}{Graphs}& 188 & 344& 1,113 & 1,000 & 1,500 & 5,000 \\
    \multicolumn{2}{c|}{Classes}& 2 & 2 & 2 & 2 & 3 & 3 \\
    \multicolumn{2}{c|}{ Avg. \# Nodes}& 17.9 & 25.5 & 39.1 & 19.8 & 13.0 & 74.5 \\
    \multicolumn{2}{c|}{ Max. \# Nodes}& 28 & 109 & 620 & 136 & 89 & 492  \\
  \hline
   \hline
    \end{tabular}
\end{table*}

\subsection{Comparison Results}

Table~\ref{tab:result} summarizes the graph classification results of the proposed UGCNs. 
We can observe that:
(1) The proposed S-UGCN and G-UGCN outperform GCN which indicates that
the learnable depthwise convolution weights perform better than the fixed one.
Moreover, S-UGCN and USGCN also perform better than GAT which reflects the importance of extracting different channel information through specific convolution filters.
(2) The proposed methods outperform most state-of-the-art GNNs.
In particular, on dataset MUTAG and PTC-MR, S-UGCN exceeds the competitive model DropGIN~\cite{dropgnn} by 5.16\% and 2.82\% respectively and G-UGCN exceeds DropGIN by 4.04\% and 2.23\% on this two datasets respectively.
Meanwhile, our methods keep better performance on other datasets.
S-UGCN obtains the best accuracy 76.58\% on PROTEINS.
G-UGCN obtains the best accuracy 76.50\%, 54.07\% and 81.84\% on three social datasets respectively.
These results indicate that the proposed models can better encode the discriminate information of different feature channels to obtain richer graph representations.
(3) Our methods perform better than some other graph representation methods.
Especially on MUTAG and PTC-MR datasets, S-UGCN exceeds the competitive model CMRL~\cite{cuco} by 5.86\% and 6.62\% respectively and
G-UGCN exceeds CMRL~\cite{cuco} by 4.74\% and 6.03\% respectively.
It further demonstrates the superiority of our methods on addressing graph classification task.
(4) Our S-UGCN and G-UGCN significantly outperform traditional graph kernel based methods.
%
\begin{table*}[!ht]
    \centering
    \caption{Classification accuracy on three bioinformatics datasets and three social datasets.
    The color of \textcolor{red}{red} and \textcolor{blue}{blue} denote the best and second performance, respectively.}
    \label{tab:result}
    \begin{tabular}{c|c|c|c|c|c|c|c}
    \hline
    \hline
        \multicolumn{2}{c|}{Methods} & MUTAG & PTC-MR & PROTEINS & IMDB-BINARY & IMDB-MULTI & COLLAB  \\ \hline
        \multirow{4}{*}{Graph  Kernel} & JTQK & 85.50$\pm$0.55 & 58.50$\pm$0.39 & 72.86$\pm$0.41 & 72.45$\pm$0.81 & 50.33$\pm$0.49 & 76.85$\pm$0.40 \\
        \cline{2-8}
        ~ & WL-Subtree & 82.05$\pm$0.36 & 58.26$\pm$0.47 & 73.52$\pm$0.43 & 71.88$\pm$0.77 & 49.50$\pm$0.49 & 77.39$\pm$0.35  \\ \cline{2-8}
        ~ & CORE SP & 88.29$\pm$1.55 & 59.06$\pm$0.93 & - & 72.62$\pm$0.59 & 49.43$\pm$0.42 & -  \\ \cline{2-8}
        ~ & GK & 83.50$\pm$0.60 & 58.65$\pm$0.40 & 71.67$\pm$0.55 & 65.87$\pm$0.98 & 45.42$\pm$0.87 & 72.83$\pm$0.28  \\ \hline \hline

        \multirow{6}{*}{Graph Repersentation} & DiffPool & 82.66 & - & 76.25 & - & - & 75.48  \\ \cline{2-8}
        ~ & GMTPool & 83.44$\pm$1.33 & - & 75.09$\pm$0.59 & 73.48$\pm$0.76 & 50.66$\pm$0.82 & 80.74$\pm$0.54  \\ \cline{2-8}
        ~ & CMRL & 89.70$\pm$1.10 & 62.50$\pm$1.70 & - & 74.20$\pm$0.70 & 51.20$\pm$0.50 & -  \\ \cline{2-8}
        ~ & InfoGraph & 89.01$\pm$1.13 & 61.65$\pm$1.43 & - & 73.03$\pm$0.87 & 49.69$\pm$ 0.53 & -  \\ \cline{2-8}
        ~ & GraphCL & 86.80$\pm$1.34 & - & 74.39$\pm$0.45 & - & - & 71.36$\pm$1.15  \\ \cline{2-8}
        ~ & CUCO & 90.55$\pm$0.98 & - & 75.91$\pm$0.55 & - & - & 72.30$\pm$0.34  \\ \hline\hline

        \multirow{10}{*}{GNNs} & DGCNN & 85.83$\pm$1.66 & 58.59$\pm$2.47 & 75.54$\pm$0.94 & 70.03$\pm$0.86 & 47.83$\pm$0.85 & 73.76$\pm$0.49  \\ \cline{2-8}
        ~ & PSCN & 88.95$\pm$4.37 & 62.29$\pm$5.68 & 75.00$\pm$2.51 & 71.00$\pm$2.29 & 45.23$\pm$2.84 & 72.60$\pm$2.15  \\ \cline{2-8}
        ~ & GCCNN & - & 66.01$\pm$5.91 & \textcolor{blue}{76.40$\pm$4.17} & 71.69$\pm$3.40 & 48.50$\pm$4.10 & 77.71$\pm$2.51  \\ \cline{2-8}
        ~ & GIN & 89.40$\pm$5.60 & 64.60$\pm$7.00 & 76.20$\pm$2.80 & 75.10$\pm$5.10 & \textcolor{blue}{52.30$\pm$2.80} & 80.20$\pm$1.90  \\ \cline{2-8}
        ~ & BASGCN & 90.04$\pm$0.82 & 60.50$\pm$0.77 & 76.05$\pm$0.57 & 74.00$\pm$0.87 & 50.43$\pm$0.77 & 79.60$\pm$0.83  \\ \cline{2-8}
        ~ & DAGCN & 87.22$\pm$6.10 & 62.88$\pm$9.61 & 76.33$\pm$4.30 & - & - & -  \\ \cline{2-8}
        ~ & DropGIN & 90.40$\pm$7.00 & 66.30$\pm$8.60 & 76.30$\pm$6.10 & 75.70$\pm$4.20 & 51.40$\pm$2.80 & -  \\ \cline{2-8}
        ~ & FactorGCN & 89.90$\pm$6.50 & - & - & 75.30$\pm$2.70 & - & 81.20$\pm$1.40  \\ \cline{2-8}
        ~ & WEGL & - & 66.20$\pm$6.90 & 76.30$\pm$3.90 & 75.20$\pm$5.00 & 52.30$\pm$2.90 & 80.60$\pm$2.00  \\ \cline{2-8}
        ~ & GAT & 88.67$\pm$7.93 & 62.65$\pm$8.53 & 73.42$\pm$4.86 & 73.10$\pm$3.33 & 51.27$\pm$3.35 & 77.56$\pm$1.33  \\ \cline{2-8}
        ~ & GCN & 90.56$\pm$7.78 & 62.65$\pm$8.93 & 72.70$\pm$1.89 & 73.50$\pm$4.01 & 51.60$\pm$2.44 & 80.23$\pm$1.12  \\ \hline\hline

        \multirow{2}{*}{Ours} & S-UGCN & \textcolor{red}{95.56$\pm$4.16} & \textcolor{red}{69.12$\pm$6.61} & \textcolor{red}{76.58$\pm$3.32} & \textcolor{blue}{76.30$\pm$2.76} & 52.27$\pm$3.70 & \textcolor{blue}{81.50$\pm$1.21} \\ \cline{2-8}
         & G-UGCN & \textcolor{blue}{94.44$\pm$2.48} & \textcolor{blue}{68.53$\pm$8.22} & 75.50$\pm$3.44 & \textcolor{red}{76.50$\pm$3.07} & \textcolor{red}{54.07$\pm$4.77} & \textcolor{red}{81.84$\pm$1.48}  \\
         \hline\hline
    \end{tabular}
\end{table*}

\subsection{Model Analysis}

\subsubsection{Capacity Analysis}

\begin{figure}[!htbp]
	\centering
\includegraphics[width=0.42\textwidth]{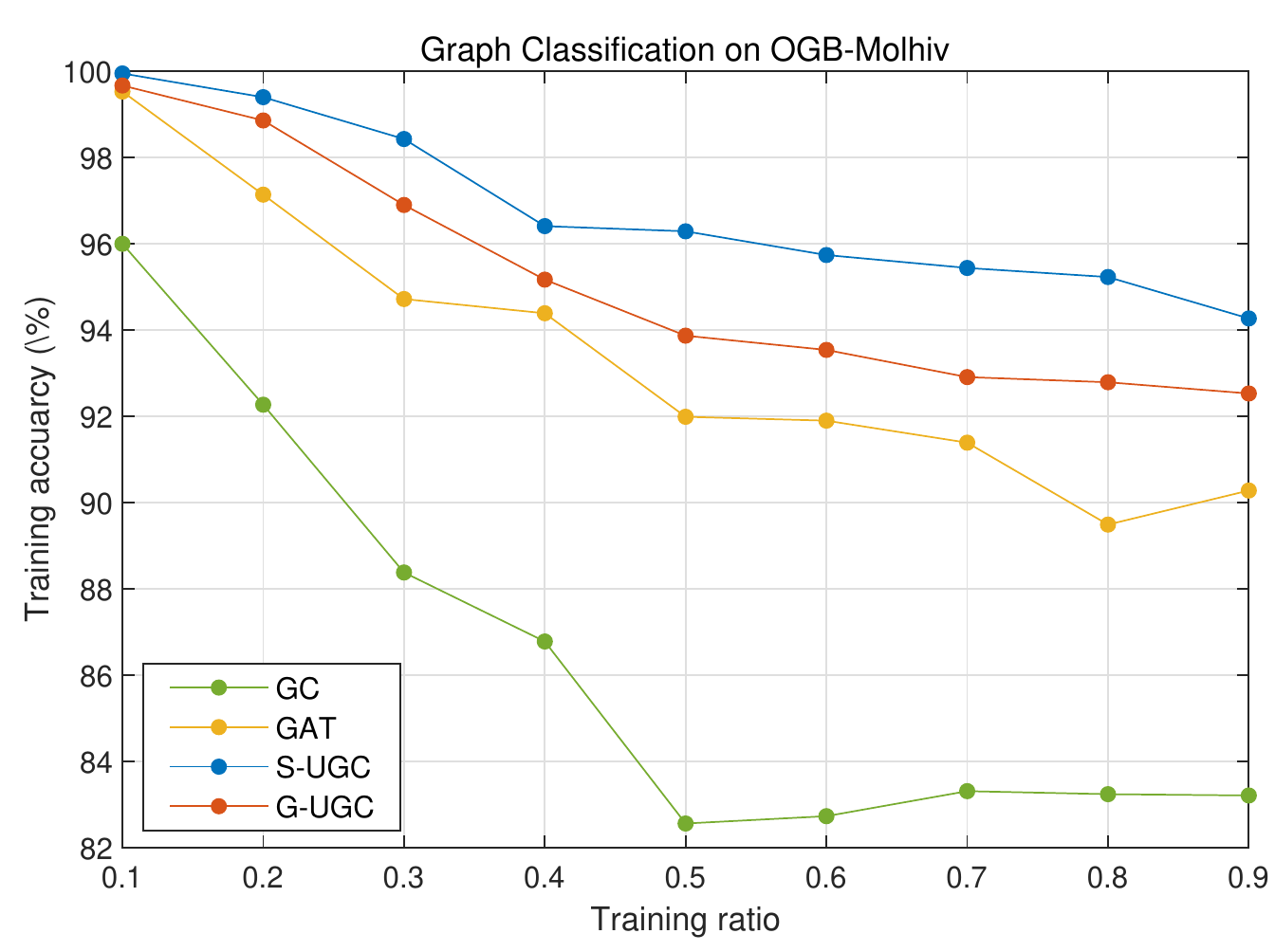}
	\caption{The comparison of network capacity on OGB-Molhiv dataset.}\label{fig:capacity-graph}
\end{figure}

For a large dataset, a neural network with higher capacity can achieve higher training accuracy.
We employ a larger dataset OGB-Molhiv~\cite{ogb} to test the ability of different GCNs to fit a large amount of training data. 
OGB-Molhiv~\cite{ogb} dataset contains 41,127 graphs which can be divided into 2 categories.
For fair comparison, all compared GCNs adopt two blocks without skip-connection.
Each block has the same architecture shown in Figure \ref{fig:block}.
The hidden units is set to $32$.
The learning rate and  number of epoch are set to $0.005$ and $500$, respectively.
Figure \ref{fig:capacity-graph} reports the training accuracy of GCN~\cite{gcn}, GAT\cite{gat}, S-UGCN and G-UGCN under the training ratio ranging from $0.1$ to $0.9$ on this dataset.
Here, we can observe that our methods maintain the obviously higher training accuracy when compared to GCN~\cite{gcn} and GAT~\cite{gat}.
Especially for GAT and S-UGCV, these two models have the same amount of learning parameters but we can note that S-UGCV obviously outperforms classic GAT model~\cite{gat}.
These clearly indicate the higher network capacity of the proposed UGCNs on conducting complex graph data representation and learning.   
Also, S-UGCN shows higher capacity than G-UGCN in terms of training accuracy. 
%

\subsubsection{Generalization Ability Analysis}

\begin{figure*}[tbp]
	\centering
\includegraphics[width=1.0\textwidth]{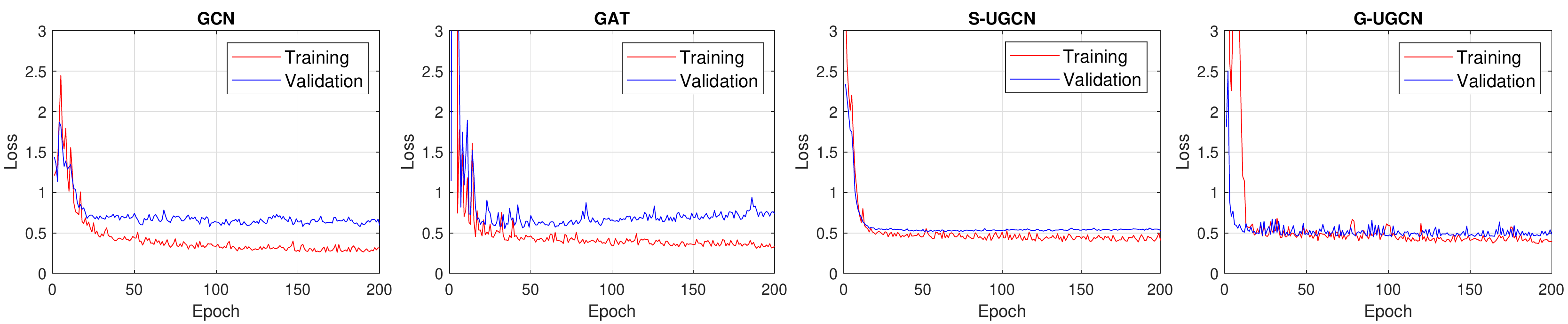}
	\caption{The comparison of generalization ability on IMDB-MULTI dataset.}\label{fig:generalization}
\end{figure*}

Figure \ref{fig:generalization} compares the training loss and validation loss of GCN~\cite{gcn}, GAT\cite{gat}, S-UGCN and G-UGCN on IMDB-MULTI dataset~\cite{datasets}.
For fair comparison, these models utilize the same architecture and parameter setting.
The gap between the training loss and validation loss can reflect the generalization ability of network model.
Here we can see that, our S-UGCN and G-UGCN have smaller gap compared to GCN~\cite{gcn} and GAT\cite{gat}.
It indicates the stronger generalization ability of the proposed methods.
Besides, compared to S-UGCN, G-UGCN shows better performance which further suggests the superiority of G-UGCN in terms of generalization.

\section{Conclusion}

This paper shows the deep connection between existing GCNs and classic CNNs from the perspective of depthwise separable convolution.
We show that both GCN and GAT indeed perform some specific depthwise separable convolution operations.
This connection analysis can enable us to design some new high capacity GCNs flexibly by  borrowing the designing sprits of CNNs.
As showcases, in this work, we develop Unified GCNs (UGCNs) by following the architecture of MobileNet and standard CNN.
Promising experiments on several graph representation benchmarks demonstrate the effectiveness and advantages of the proposed UGCNs.

\bibliography{iclr2022_conference}
\bibliographystyle{ieeetr}
\ifCLASSOPTIONcaptionsoff
  \newpage
\fi

\end{document}